\documentclass[sigconf]{acmart}

\usepackage[whole]{bxcjkjatype}  
\usepackage{booktabs}
\usepackage{multirow}
\usepackage{mathtools}
\usepackage{siunitx}
\usepackage{amsfonts}

\usepackage{comment}
\usepackage{algorithm}
\usepackage{algpseudocode}
\usepackage[capitalize]{cleveref}
\crefname{section}{Sec.}{Secs.}
\Crefname{section}{Section}{Sections}
\Crefname{table}{Table}{Tables}
\crefname{table}{Tab.}{Tabs.}

\newcommand{\nishi}[1]{\textcolor{cyan}{[\textsc{NISHI:} #1]}}

\AtBeginDocument{%
  \providecommand\BibTeX{{%
    \normalfont B\kern-0.5em{\scshape i\kern-0.25em b}\kern-0.8em\TeX}}}

\copyrightyear{2023}
\acmYear{2023}
\setcopyright{acmlicensed}\acmConference[MMAsia '23]{ACM Multimedia Asia 2023}{December 6--8, 2023}{Tainan, Taiwan}
\acmBooktitle{ACM Multimedia Asia 2023 (MMAsia '23), December 6--8, 2023, Tainan, Taiwan}
\acmPrice{15.00}
\acmDOI{10.1145/3595916.3626412}
\acmISBN{979-8-4007-0205-1/23/12}

\acmConference[Conference acronym 'XX]{Make sure to enter the correct
  conference title from your rights confirmation emai}{June 03--05,
  2018}{Woodstock, NY}
\acmPrice{15.00}
\acmISBN{978-1-4503-XXXX-X/18/06}

\acmSubmissionID{183}

\begin{document}

\title{Image Cropping under Design Constraints}


\author{Takumi Nishiyasu}
\email{nisiyasu@iis.u-tokyo.ac.jp}
\orcid{????}
\affiliation{%
  \institution{The University of Tokyo}
  \streetaddress{???}
  \city{Hongo}
  \state{Tokyo}
  \country{JAPAN}
  \postcode{?-?}
}

\author{Wataru Shimoda}
\email{wataru_shimoda@cyberagent.co.jp}
\affiliation{%
  \institution{Cyberagent.Inc}
  \streetaddress{???}
  \city{Shibuya}
  \state{Tokyo}
  \country{JAPAN}
  \postcode{?-?}
}

\author{Yoichi Sato}
\email{ysato@iis.u-tokyo.ac.jp}
\orcid{????}
\affiliation{%
  \institution{The University of Tokyo}
  \streetaddress{???}
  \city{Hongo}
  \state{Tokyo}
  \country{JAPAN}
  \postcode{?-?}
}

\renewcommand{\shortauthors}{Trovato and Tobin, et al.}


\begin{abstract}
Image cropping is essential in image editing for obtaining a compositionally enhanced image. 
In display media, image cropping is a prospective technique for automatically creating media content. 
However, image cropping for media contents is often required to satisfy various constraints, such as an aspect ratio and blank regions for placing texts or objects. 
We call this problem image cropping under design constraints.
To achieve image cropping under design constraints, we propose a score function-based approach, which computes scores for cropped results whether aesthetically plausible and satisfies design constraints. 
We explore two derived approaches, a proposal-based approach, and a heatmap-based approach, and we construct a dataset for evaluating the performance of the proposed approaches on image cropping under design constraints.
In experiments, we demonstrate that the proposed approaches outperform a baseline, and we observe that the proposal-based approach is better than the heatmap-based approach under the same computation cost, but the heatmap-based approach leads to better scores by increasing computation cost. 
The experimental results indicate that balancing aesthetically plausible regions and satisfying design constraints is not a trivial problem and requires sensitive balance, and both proposed approaches are reasonable alternatives.
\end{abstract}





\begin{teaserfigure}
\centering
  \includegraphics[width=1.0\textwidth]{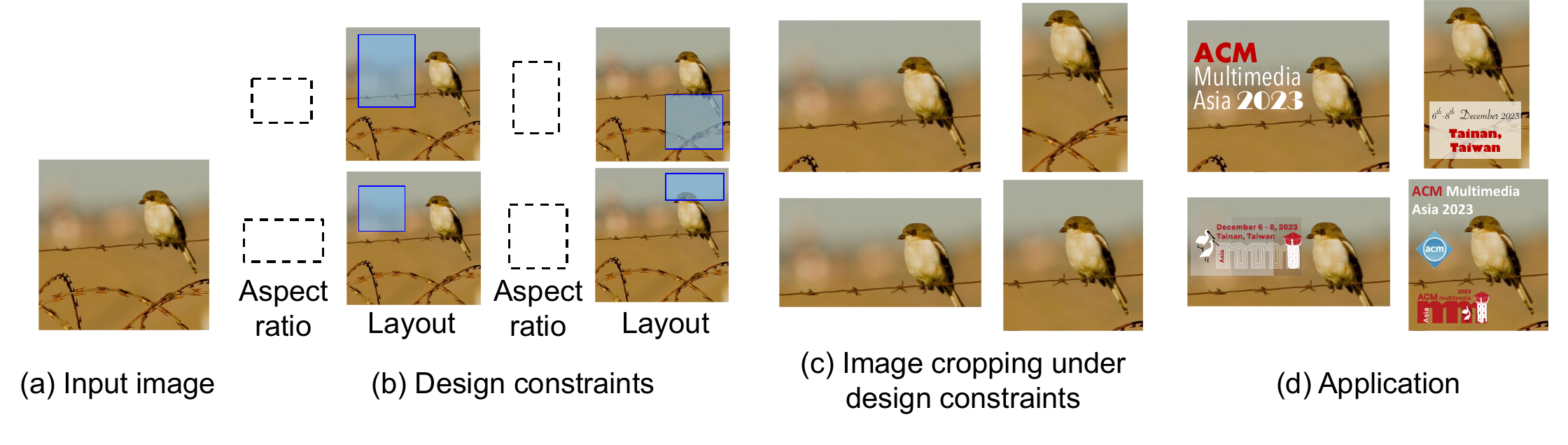}
  \vspace{-10mm}
  \caption{An Illustration of the problem setting and application. We tackle how to crop images with a given aspect ratio and a spatial layout where we want to place objects. (a) an input image, (b) design constraints, a pair of aspect ratios and layout conditions, (c) cropping results under design constraints, and (d) examples for applications.}
  \label{fig:abstract}
\end{teaserfigure}

\received{20 February 2007}
\received[revised]{12 March 2009}
\received[accepted]{5 June 2009}

\maketitle

\section{Introduction}

Image cropping is commonly used in image editing that attempts to find a better view of an image. 
In the field of display media, such as advertisements, posters, and book covers, image cropping is an indispensable technique, but still carefully done by hand.
Image cropping is often required to satisfy design constraints in content creation, for example, fixed aspect ratio for banners, specifying a blank region for placing texts, and including essential objects in an image.
In this paper, we call this problem image cropping under design constraints and tackle this challenging and practical task.

In this paper, we handle the aspect ratio and the layout condition as design constraints that are often required simultaneously in display media, and we focus on incorporating two conditions into the image cropping task. 
We show the illustration of the proposed problem setting and application in Figure.~\ref{fig:abstract}.
The aspect ratio condition ensures that the cropped image satisfies the given aspect ratio, and the layout condition defines specific regions that should be included to support the creation of empty spaces for element placement, e.g., advertisement texts or containing essential objects in cropping results.  
Some prior works~\cite{zhong2021aesthetic, li2020learning} investigate effective cropping methods for satisfying the given aspect ratio, while there is no research on layout condition-aware image cropping though it is important in the practical use of image cropping.
We explore effective approaches for satisfying both the aspect ratio condition and the layout condition.
In this paper, we propose a score function-based approach.
We define score functions that evaluate whether the result of cropping is aesthetically plausible and whether multiple conditions are satisfied and compute a total score by combining the scores.
This bottom-up approach easily extends design constraints by adding score functions for new design constraints.
As the derivative of the score function-based approach, we explore two approaches: a proposal-based approach and a heatmap-based approach.
For evaluation of the aesthetic of cropped regions, the proposal-based approach computes a forwarding pass of a neural network per candidate, i.e., computes multiple forwarding processes for an image, while the heatmap-based approach computes one forwarding pass of a neural network to obtain heatmaps.
In the heatmap-based approach, we optimize cropping region by an optimization algorithm because computing scores per region using the obtained heatmaps requires only CPU computation.
We control the search space of cropping by changing the number of proposals in the proposal-based approach, while we control the search space by changing the hyperparameters of optimization in the heatmap-based approach.

As another important contribution of this work, we prepare a dataset for evaluating image cropping under design constraints utilizing an existing dataset.
We build the new dataset by adding the design constraints that have consistency to the existing ground truth. Then, the dataset contains the set of input images with design constraints and ground truth of outputs.
We evaluate the proposed approaches using the dataset and compare the proposed approaches with a simple baseline.
We confirm that the proposed approaches outperform the baseline, and the proposal-based approach is better than the heatmap-based approach under the same computation cost, but the heatmap-based approach can achieve better scores by increasing the computation cost.
The result indicates that balancing aesthetically plausible regions and satisfying multiple conditions is not a trivial task and requires sensitive balance.

In summary, the contributions of this study are twofold: 1) We set the problem of image cropping under design constraints and propose a score function-based approach with the two derivatives for the task, and 2) We prepare a dataset for evaluation of the performance on the task of image cropping under design constraints.
\begin{figure}[t]
\centering
\includegraphics[width=\linewidth]{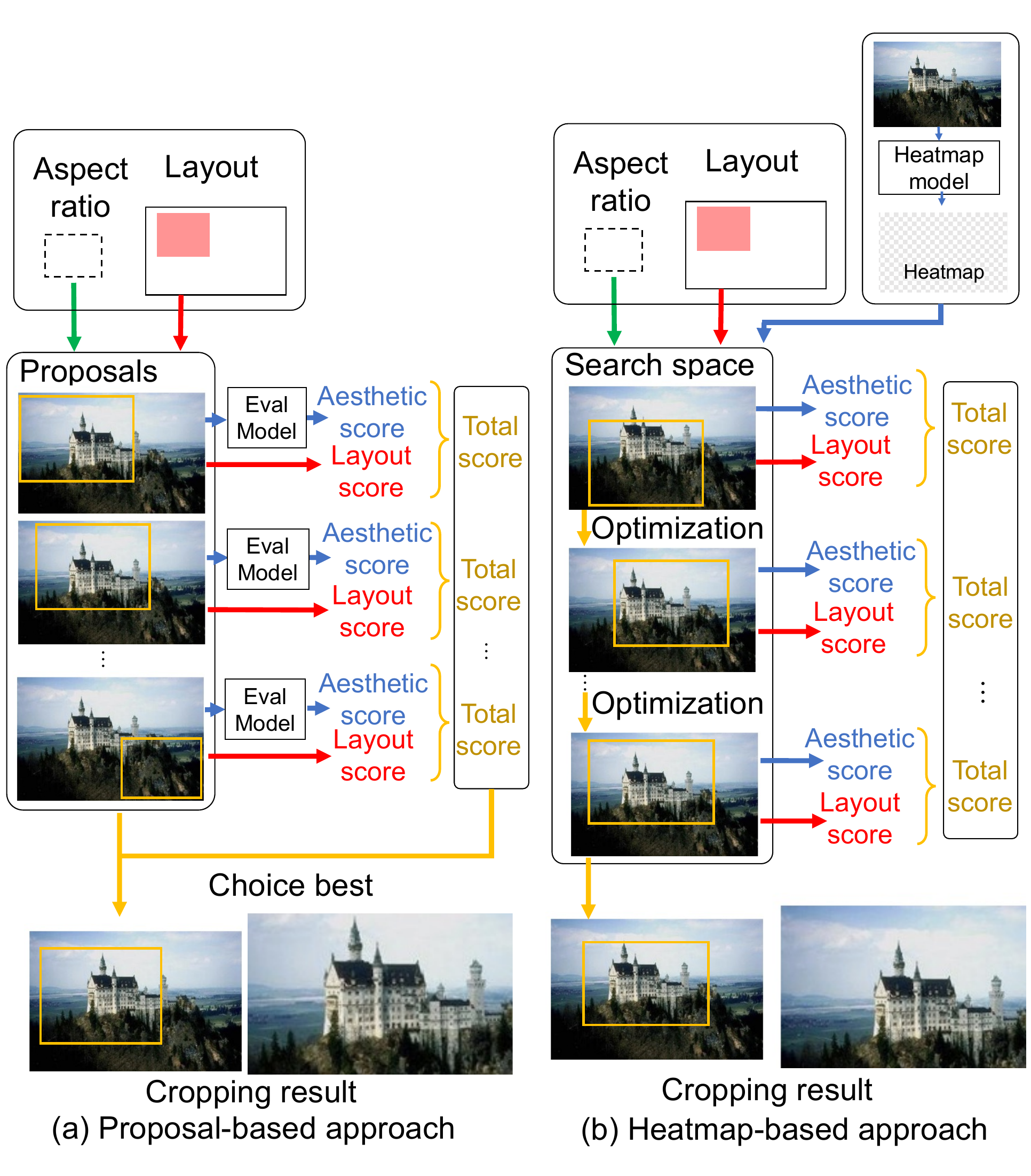}
\vspace{-5mm}
\caption{We prepare two methods: (a) a proposal-based method and (b) a heatmap-based method. (a) predicts an aesthetic score per cropping region by a scoring model. (b) generates a heatmap from an image and computes aesthetic scores using the heatmap.
}
\label{fig:model_all}
\end{figure}

\section{Related Work}

\paragraph{Image cropping}
Image cropping is a method for removing unnecessary regions for various kinds of purposes, such as improving aesthetics, obtaining thumbnails, or adjusting an aspect ratio of images.
We categorize image cropping methods into two types: exploring cropping regions directly and choosing the best region by scoring some candidate regions.


As the former approach, some prior works~\cite{cheng2010learning, ni2013learning, ardizzone2013saliency} propose to convert images to heatmaps that are related to cropping and obtain coordinates relying on the heatmaps.
Cheng et al. and Ni et al. propose to obtain heatmaps by computing edges and colors.
Ardizzone et al.~\cite{ardizzone2013saliency} propose to use saliency maps, which represent eye-catching regions.
Kao et al.~\cite{kao2017automatic} utilize a convolutional network for computing heatmaps, which tends to indicate foregrounds.
As another approach, gaze-based cropping \cite{santella2006gaze} utilizes eye-tracking information to estimate suitable cropping regions by locating important content, thumbnail cropping \cite{suh2003automatic} estimates important cropping regions from visual prominence maps.
Recently, some methods propose directly predicting coordinates using a deep neural network~\cite{chen2018cropnet,esmaeili2017fast,Jia_2022_CVPR, Hong_2021_CVPR}.


As the latter approach, there are approaches scoring proposals generated by sliding windows~\cite{fang2014automatic,chen2017learning,huang2015automatic, wang2017deep, wang2018deep}.
The scoring function~\cite{nishiyama2011aesthetic,li2018a2} for image cropping has also been explored.
Some prior works~\cite{nishiyama2009sensation,  wei2018good} attempt to obtain proposals via estimating regions that are related to cropping.
Recently, scoring pre-defined grid-based anchor boxes is a popular approach~\cite{li2019fast,murray2012ava,wei2018good,zeng2019reliable}.
In this paper, we explore both two approaches for the computation of the score function of aesthetic scores, these approaches have different computation costs for finding good optimal of the score function.

\paragraph{Conditioned image cropping}
Conditioned image cropping is an important problem for the practical use of image cropping, but there are few methods comparing image cropping.
Aspect ratio-aware image cropping is a popular task setting~\cite{zhong2021aesthetic, li2020learning}, which aims to obtain cropped results that satisfy the given aspect ratio yet are aesthetically plausible.
Recently, text description-based cropping methods get an attention~\cite{horanyi2022repurposing,zhong2022clipcrop} that extract regions corresponding to text description and maintaining aesthetics.
In this work, we focus on conditions of an aspect ratio and a layout condition.
Further, we attempt to obtain cropped regions that keep aesthetics and satisfy the given aspect ratio and specified layout, simultaneously.

\begin{figure}[t]
\centering
\includegraphics[width=1.0\linewidth]{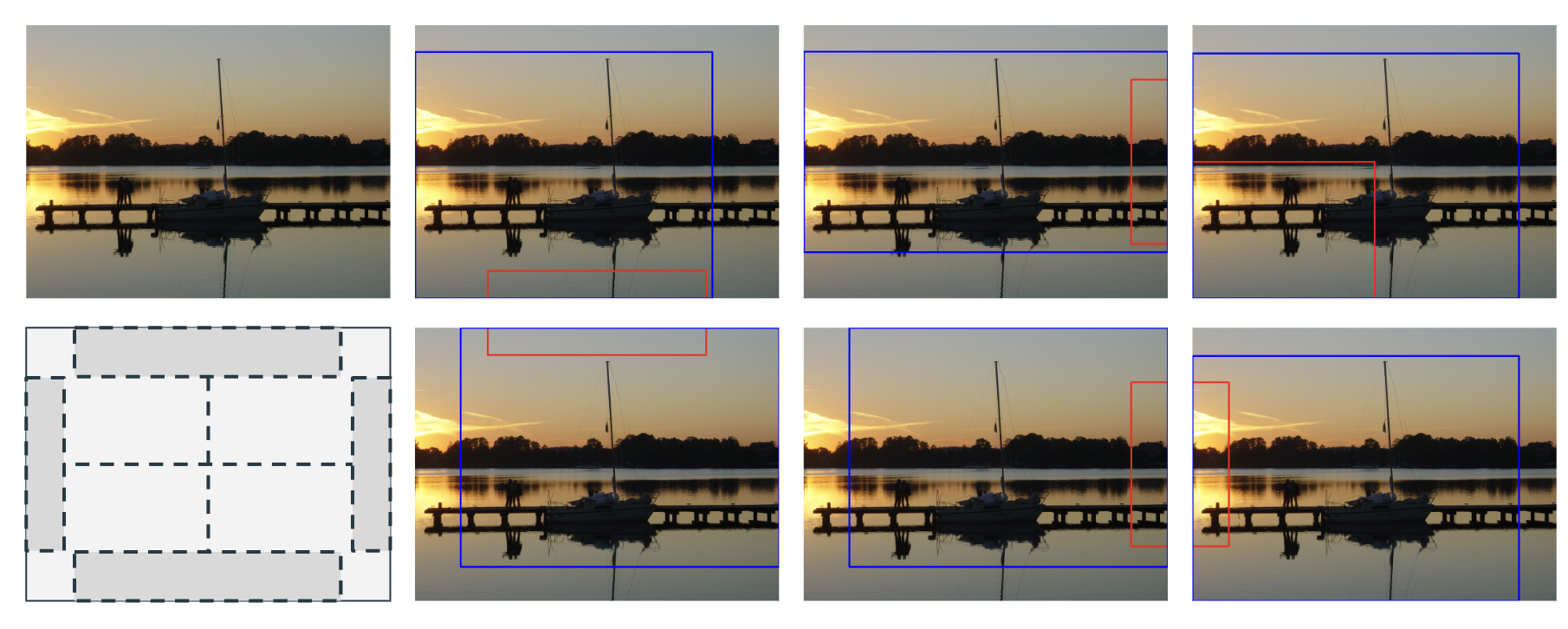}
\vspace{-5mm}
\caption{Examples of the evaluation dataset. The left image shows the original image and templates of layout conditions.
In the following images, the red boxes indicate layout conditions, while the blue boxes are the ground truth regions. 
}

\label{fig:eval_dataset}
\end{figure}
\begin{figure}[t]
\centering
\includegraphics[width=\linewidth]{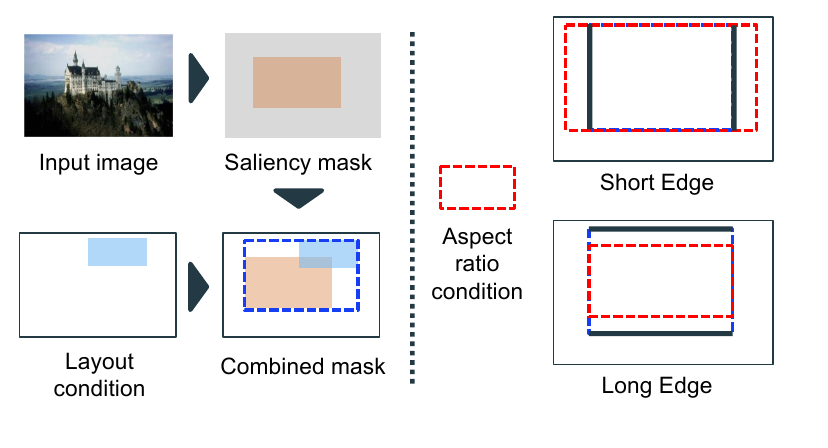}
\vspace{-5mm}
\caption{An illustration of the baseline. The baseline method creates a mask by combining a saliency mask and a layout condition. Then, the baseline adjusts the aspect ratio by reshaping the mask based on the shorter or longer edge.}
\label{fig:baseline_methods}
\end{figure}
\begin{figure*}[t]
\centering
\includegraphics[width=\linewidth]{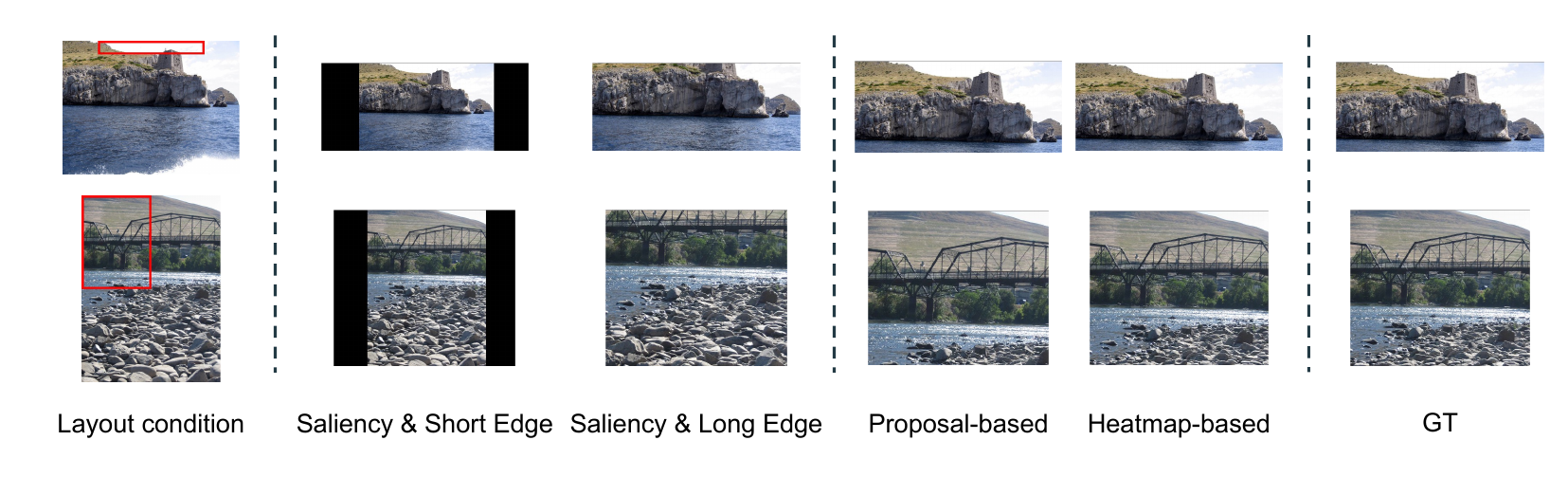}
\vspace{-8mm}
\caption{This figure shows qualitative comparisons. The left image shows an input image with layout conditions. The middle four images show the results of image cropping under design constraints by the baseline and the proposed approaches. The right figure shows the ground truths.}
\label{fig:vis_comaparison}
\end{figure*}

\section{Methodology}
We formulate the image cropping under design constraints task and propose a score function-based approach.
Further, we explore two approaches a proposal-based approach and a heatmap-based approach. 

\subsection{Problem Formulation}
Formally, we denote an input image as $x \in \mathbb{R}^{H \times W \times 3}$ and a result of cropping as $y$, where $y$ represents a bounding box region in an image and is a tuple of ($\mathrm{height},\mathrm{width},\mathrm{position}_{\mathbf{X}},\mathrm{position}_{\mathbf{Y}}$).
In this paper, we handle two types of conditions: an aspect ratio $\omega$ and a layout condition $\varphi$, which represents the region that should be included in the result of cropping.
Our objective is that a result of cropping $y$ is aesthetically improved from the original input image while satisfying the two conditions. 

We propose a score function-based approach to address satisfying multiple conditions in image cropping.
Let $V_\textrm{aesth}(x|y)$ be the score function for an aesthetic of the image $x$ cropped with the region $y$, $V_\textrm{aspect}(\omega|y)$ be the score function for the aspect ratio of the region $y$, and $V_\textrm{layout}(\varphi|y)$ be the score function for the specific layout against the region $y$.
We approximate that the score functions have linear relationships and define a comprehensive score function $V(x, \omega, \varphi | y)$, which is given by:

\begin{align}
   V(x, \omega, \varphi | y) = V_\textrm{aesth}(x|y)+\lambda_{1} V_\textrm{aspect}(\omega|y)+\lambda_{2} V_\textrm{layout}(\varphi|y), 
   \label{eq:_score}
\end{align}
We use this score function in Eq.~\ref{eq:_score}.
where $\lambda_{1}$ and $\lambda_{2}$ are hyperparameters for balancing each score function.
To simplify, we regard the condition of an aspect ratio should be completely satisfied by making the $\lambda_{1}$ sufficiently large.
The re-formulated score function is given by :
\begin{align}
V(x, \varphi | y) = V_\mathrm{aesth}(x|y)+\alpha V_\mathrm{layout}(\varphi|y), 
   \label{eq:score}
\end{align}
where $\alpha$ is a hyperparameter.
We represent the score function for the layout condition by computing the recall in the cropped region $y$ and the specified region.
We define the score function for this layout condition as:

\begin{align}
   V_\mathrm{layout}(\varphi | y) = \frac{\left|\mathcal{B}_{y} \cap \mathcal{B}_{\varphi}\right|}{\left|\mathcal{B}_{\varphi}\right|}, 
   \label{eq:layout_score}
\end{align}
where $\mathcal{B}_{y}$ is a pixel set inside the region $y$ and $\mathcal{B}_{\varphi}$ is a set of pixels for the specified layout $\varphi$, respectively. 

We assume that the higher score of the score function reflects better results.
We regard image cropping under design constraints as an optimization problem of this score function.
Let $\mathcal{Y}_{\omega}$ be a set for search space for the result of cropping $y$, which satisfies the given aspect ratio condition $\omega$.
We solve the optimization problem $\Psi$ against the score function $V$, image, and conditions in the search space $\mathcal{Y}_{\omega}$.
We denote this optimization process as the following:
\begin{align}
y = \Psi(V, x, \varphi,\mathcal{Y}_{\omega}).   
   \label{eq:optmization}
\end{align}
There are some approaches for computing aesthetic score $V_\mathrm{aesth}(x|y)$ against $y$ and how to solve the optimization problem. We explore two approaches the heatmap-based approach and the proposal-based approach for this problem.

\subsection{Proposal-based Approach}
In a proposal-based approach, we prepare candidates' regions that are bounding boxes and compute scores via score functions per candidate. Then, we explore the best region for image cropping from the candidates based on the scores.
We illustrate a proposal-based approach in Figure.~\ref{fig:model_all}(a).
We denote the aesthetic score as $f^p(x|y)$ and use it for the score function $V_\mathrm{aesth}^{p}(x|y)$:
\begin{align}
   V_\mathrm{aesth}^{p}(x|y) = f^p(x|y). \label{eq:heatmap}
\end{align}
We use a pre-trained model GAICv2~\cite{zeng2020cropping} for computing the proposal-based aesthetics scores.

We use this score function for optimization in Eq.~\ref{eq:score}.
We prepare grid-based bounding boxes $\mathcal{Y}^{p}_{\omega}$, which includes only the proposals that satisfy the given condition of aspect ratio $\omega$.
Unlike the prior work~\cite{zeng2020cropping}, we increase the number of proposals to address various types of conditions.
Specifically, we define a minimum size of height and width and linearly increase the sizes as far as possible against an input image. For each set of height and width, we generate proposals with a fixed offset, which is the same as the minimum size of height and width. We double the size of the offset for an extreme aspect ratio whose offset is too small.
The total score function in the proposal-based approach is given by:
\begin{align}
V^{p}(x|y, \varphi) = V_\mathrm{aesth}^{p}(x|y)+\alpha V_\mathrm{layout}(\varphi|y)
   \label{eq:optmimal_prop}
\end{align}

We define an optimization problem $\Psi^{p}$ using the score function against the proposals $\mathcal{Y}^{p}_{\omega}$ by:
\begin{align}
   y = \Psi^{p}(V^{p}, x, \varphi, \mathcal{Y}^{p}_{\omega}).
   \label{eq:opt_heatmap}
\end{align}
The search space $\mathcal{Y}^{p}_{\omega}$ is not so large, and we solve this optimization problem by exhaustive search.



\subsection{Heatmap-based Approach}
\label{sec:heatmap}
The heatmap-based approach extracts aesthetic information via a deep neural network as heatmaps and computes aesthetic scores for each region by a simple process of the heatmap. 
We illustrate this approach in Figure.~\ref{fig:model_all}(b). 

Following the prior work~\cite{zhang2022human}, we convert multiple candidate regions that designers annotate to heatmaps by averaging the multiple cropping candidate regions.
We train a neural network to predict the heatmap. We assume that the heatmap includes sufficient aesthetic information for image cropping and evaluate each cropping result without a repeat of neural network computation using the heatmap.
Let a heatmap $z \in \mathbb{R}^{H \times W}$ against an input image $x$ be: 
\begin{align}
   z = f^{h}(x).
   \label{eq:heatmap}
\end{align}

We use the heatmap for computing an aesthetic score $V_\mathrm{aesth}^{h}(x|y)$ for the result of cropping $y$ by: 
\begin{align}
   V_\mathrm{aesth}^{h}(x|y) & = V_{\mathrm{RoI}}(z|y) + V_{\mathrm{RoD}}(z|y), \\
   V_{\mathrm{RoI}}(z|y) & =\sum_{b \in \mathcal{B}_{y}}z(b), \\
   V_{\mathrm{RoD}}(z|y) & = \sum_{b \notin \mathcal{B}_{y}} 1-z(b), 
   \label{eq:heatmap_score}
\end{align}
where $\mathcal{B}_{y}$ is a pixel set inside the bounding box $y$. 
Higher values of the score function $V_\mathrm{aesth}^{h}(x|y)$ means that the total amount of the heatmap is large in the region $y$ and small in the outside of the region $y$.
We train the model following the prior work~\cite{zhang2022human}. 

We use this score function in Eq.~\ref{eq:score}.
The score function in the heatmap-based approach is given by :
\begin{align}
V^{h}(x|y, \varphi) = V_\mathrm{aesth}^h(x|y)+\alpha V_\mathrm{layout}(\varphi|y), 
   \label{eq:optmimal_heatmap}
\end{align}
We find optimal $y$ in this score function by solving a black-box optimization problem.
We define a search space $\mathcal{Y}^{h}_{\omega}$, which consists of an optimization function for the heatmap approach $\Psi^{h}$ as:

\begin{align}
   y = \Psi^{h}(V^{h}, x, \varphi, \mathcal{Y}^{h}_{\omega}).
   \label{eq:opt_heatmap}
\end{align}
In this paper, we use the optimization library Optuna~\cite{optuna_2019} for the solver of the optimization problem.

\begin{figure*}[t]
\centering
\includegraphics[width=\linewidth]{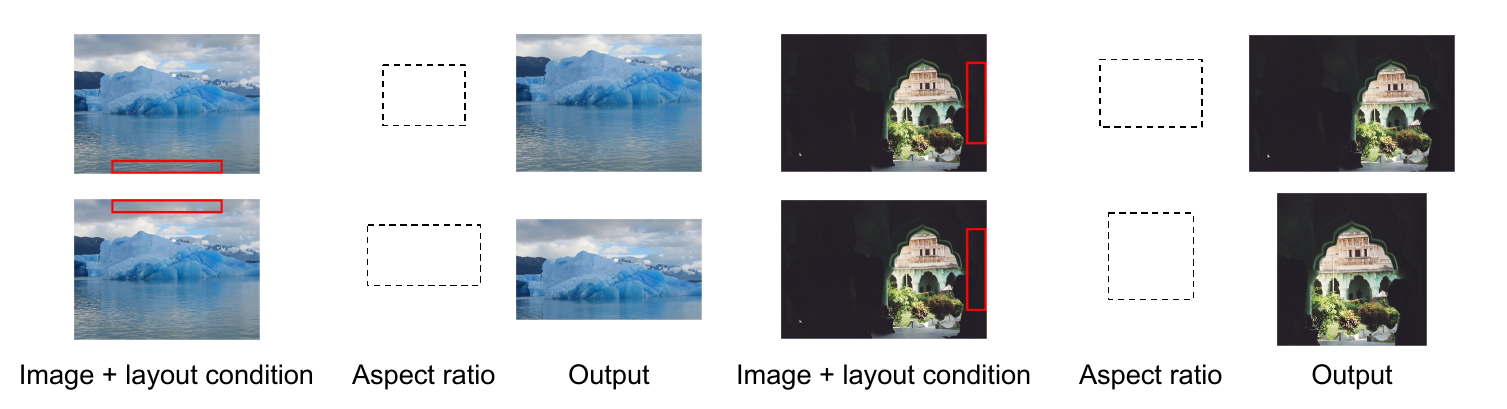}
\vspace{-5mm}
\caption{Examples of the image with multiple conditions and the results of cropping by the heatmap-based approach.}
\label{fig:several_results_examples}
\end{figure*}

\section{Dataset}






To evaluate image cropping models under design constraints, we prepared a new dataset based on the existing dataset for aesthetic image cropping\cite{fang2014automatic}, with additional design constraints applied.
We use an existing dataset, FLMS~\cite{fang2014automatic}, which consists of 500 images with 10 bounding boxes given by experts corresponding to aesthetic region.
We add design constraints that have consistency with the ground truth data. Then, the dataset contains the set of input images with design constraints and ground truth of outputs.
To be concrete, we prepare templates of layout conditions and check the consistency of the layout conditions with 10 ground truth boxes. Then, we only pick the sets of the layout condition and the ground truth that have consistency.
The definition of consistency is that whether the ground truth includes the layout condition or not.

As per the template layout conditions, we prepare eight types of bounding boxes.
To be concrete, we place four narrow boxes along each image side and four large boxes by dividing images from a center point with vertical and horizontal lines.
We expect that these blanks are useful for the placement of something like text elements or logos.
Further, we add aspect ratio conditions by computing aspect ratios from the bounding box of ground truth. Then we obtain the set of input images with design constraints and ground truth of outputs.
For each pair, when the ground truth region encompasses a layout pattern, we simply retain the aspect ratio of the ground truth region as an input condition.
Through this process, we achieve a set comprising the image, the layout pattern, the aspect ratio, and the ground truth region.

Figure.~\ref{fig:eval_dataset} shows the example of the evaluation dataset: images, layout conditions, and corresponding crops.
Red boxes in images represent the specified layout conditions while blue boxes are ground truth crops that contain the red regions, i.e., satisfies layout conditions.
The first, second, and fourth examples show examples of narrow boxes in Figure.~\ref{fig:eval_dataset}.
Totally, we collect 4426 sets of design constraints and corresponding crops.
If models can handle these two conditions, they can get good results of cropping by specifying objects that should be or not be included in cropping results with a given aspect ratio.

\section{Experiments}
In experiments, we evaluate the effectiveness of the proposed approaches on the benchmark and analyze the effect of hyperparameters of the proposed approaches on the performance. We also show examples of the practical application of the proposed approaches under design constraints.

\subsection{Settings}

\paragraph{Baseline}
We introduce a baseline method for image cropping under design constraints.
The baseline is for investigating whether rough aesthetic evaluation and satisfying conditions are enough for image cropping under design constraints.
As a rough aesthetic evaluation method, we use saliency maps~\cite{lou2022transalnet} according to the success of some prior works~\cite{ardizzone2013saliency,marchesotti2009framework}.
The baseline creates a mask combining a saliency and a layout mask and reshapes the mask with simple modifications to satisfy two conditions.
The threshold of saliency masks is $0.01$.
We call the baseline of two derivatives \textbf{Saliency \& Short Edge} and \textbf{Saliency \& Long Edge} shown in Figure.~\ref{fig:baseline_methods}.
\textbf{Saliency \& Short Edge} re-frames the mask to adjust to the given aspect ratio, taking the shorter side of the bounding box as the reference for fitting the mask.
\textbf{Saliency \& Long Edge} re-frames the mask to adjust to the given aspect ratio, taking the longer side of the bounding box as the reference for fitting the mask.




\paragraph{Evaluation metric}
As an evaluation metric, we use the IoU, a standard metric for evaluating bounding box-based tasks ~\cite{Hong_2021_CVPR,wang2017deep,wang2018deep}. Unlike previous works~\cite{zeng2019reliable, zeng2020cropping,Pan_2021_ICCV} that use a ranking-style evaluation, we use the IoU for evaluating bounding box-based tasks like  ~\cite{Hong_2021_CVPR,wang2017deep,wang2018deep}. This is because we do not focus on proposing multiple candidates in this paper. 

\paragraph{Implementation details}
The feature extractor of the heatmap-based approach is VGG16~\cite{simonyan2014very}. The model is trained using the Adam optimizer~\cite{kingma2014adam} with weight decay of 1e−4. The initial learning rate is set to 3.5e−4 and decays by $\times 0.1$ every 5 epochs, totally 30 epochs. We used CPCDatset~\cite{wei2018good}, which has a large variation in the aspect ratio of the bounding boxes. 
For the backbone of the proposal-based approach, we use a pre-trained model of GAICv2 ~\cite{zeng2020cropping}. 
We empirically set the hyperparameter $\alpha$ to 1e+4 in Eq.~\ref{eq:score}.
All models are tested with a single NVIDIA Tesla T4. 
Further details about the algorithm of optimization, bounding box generation, and model architecture are provided in \textbf{supplementals}.

\subsection{Results}

\begin{table*}[tbh]
\begin{tabular}{cc}
\begin{minipage}{.3\textwidth} 
        \begin{minipage}[h]{\textwidth}
            \begin{center}
            \caption{Comparisons on the FLMS dataset.\label{tab:multi_condition}}
                \begin{tabular}{@{}l|c@{}}
                    \toprule
                    Methods  & IoU↑ \\ \midrule
                    Saliency \& Short Edge  &  0.7134   \\ 
                    Saliency \& Long Edge &  0.7563   \\ \midrule
                    Proposal-based   &0.8106 \\
                    Heatmap-based    &\textbf{0.8450}   \\ \bottomrule
                \end{tabular}
            \end{center}
        \end{minipage}
    \\ 
    \end{minipage}
    
    
  
  \hspace{1.5mm}
  
    \begin{minipage}[h]{.3\textwidth}
    \begin{center}
    \caption{The relationships in the IoU score with computation cost. $\dagger$ and $\ddag$ mean increasing large and small boxes, respectively.   \label{tab:hyperparameters_proposal}}
    \begin{tabular}{cccc}
        \toprule
        \# proposal  & IoU↑   & time[s]\\ \midrule
         145.59   & 0.3912 & 0.2620\\ 
         $\dagger$281.27  & 0.6357 & 0.5052\\ 
         $\dagger$303.03   & 0.8106 & 0.5452\\
         $\ddag$517.93   & 0.8096 & 0.8822\\
         $\ddag$ 801.33   & 0.8088 & 1.4042\\ 
        \bottomrule
    \end{tabular}
    \end{center}
  \end{minipage}
  
  \hspace{1.5mm}

    \begin{minipage}[h]{.3\textwidth}
    \begin{center}
    \caption{The relationships in the iteration of optimizaiton and IoU score with computaiton cost.  \label{tab:optimization_param2}}
    \begin{tabular}{cccc}
        \toprule
        iteration & IoU↑   & time[s]\\ \midrule
        10        & 0.4300 & 0.1697\\
        20        & 0.6588 & 0.4267\\
        50        & 0.7895 & 1.2415\\
        100       & 0.8283 & 2.7243\\
        200       & 0.8391 & 5.9069\\
        500       & \textbf{0.8450} & 19.950\\ \bottomrule
    \end{tabular}
    \end{center}
  \end{minipage}


\end{tabular}
\end{table*}

\paragraph{Comparisons}
Table.~\ref{tab:multi_condition} and Figure.~\ref{fig:vis_comaparison} shows the comparisons of image cropping under design constraints.
We observe that the proposed approaches outperform the baseline quantitatively and qualitatively. 
The baseline tends to lose aesthetics for satisfying a given aspect ratio and a specific layout while our approaches find a good view in the areas that satisfy given conditions. 
The heatmap-based approach shows the better score than the proposal-based approach.
These results indicate that the score functions are effective for image cropping under design constraints, and the heatmap-based approach can achieve better optimal comparing the proposal-based approach.
Figure.~\ref{fig:several_results_examples} shows examples of the heatmap approach that demonstrate the proposed approach picks aesthetically good views corresponding to the given conditions.

\paragraph{Effect of Optimization Parameters and Computation Cost}

We show the trade-off for the proposal-based approach in Table.~\ref{tab:hyperparameters_proposal} and the heatmap-based approach in Table.~\ref{tab:optimization_param2}, respectively.
In the proposal-based approach, we control the number of proposals by changing the size of the offset and the sliding window.
Table.~\ref{tab:hyperparameters_proposal} shows the performance of the proposal-based approach saturates the performance against increasing the number of proposals. Especially increasing small bounding boxes for proposals tends to lead to worse results. 
The results of Table.~\ref{tab:optimization_param2} indicates that the heatmap-based approach has the trade-off of the performance and computation cost in the iteration of optimization, and fine fitting leads to large improvement comparing the proposal-based approach, though it requires a high computation cost, e.g., iteration 500 takes 20 seconds.
The proposal-based approach achieves better performance than the heatmap-based approach under the same computation cost, while the heatmap-based approach achieves better performance with more computation cost.
This result indicates that balancing aesthetically plausible areas and satisfying multiple conditions is not a trivial task and requires sensitive balance, and both proposed approaches are reasonable alternatives.

\paragraph{Performance of Unconditional Image Cropping}
Though our focus is image cropping under design constraints, we show the performance of proposed approaches for the aesthetic on the popular image cropping benchmarks.
We obtain cropping results of the score functions for the aesthetic by ignoring conditions in Eq.~\textcolor{red}{1}.
Table.\ref{tab:eval_IoU} shows the comparisons of the IoU scores for state-of-the-art methods and score functions in proposed approaches on FCDB ~\cite{chen2017quantitative} and FLMS~\cite{fang2014automatic}. 
We observe that the both proposal-based approach and heatmap-based aproach also show comparable performance against state-of-the-art methods.

\begin{table}[tb]
\centering
\caption{Comparison with state-of-the-art methods on FCDB and FLMS. The evaluation metric is the IoU ↑.\label{tab:eval_IoU}}
\begin{tabular}{l|cc}
\toprule
model  & FCDB~\cite{chen2017quantitative} & FLMS~\cite{fang2014automatic}\\ \midrule
Fang et al. ~\cite{fang2014automatic} &  -         & 0.740\\
DIC~\cite{wang2018deep}    & 0.650          & 0.810          \\
A2-RL~\cite{li2018a2}    & 0.664          & 0.821          \\
VPN~\cite{wei2018good}     & 0.711          &  0.835              \\
GAICv1~\cite{zeng2019reliable} & 0.674               & 0.834               \\
CACNET~\cite{Hong_2021_CVPR}   & 0.718 & 0.854 \\ \midrule

Proposal-based(GAICv2~\cite{zeng2020cropping})  & 0.673          &  0.836         \\
Heatmap-based        & 0.658   & 0.825  \\\bottomrule
\end{tabular}
\end{table}

 \begin{figure}[hbt]
\centering
\includegraphics[width=\linewidth]{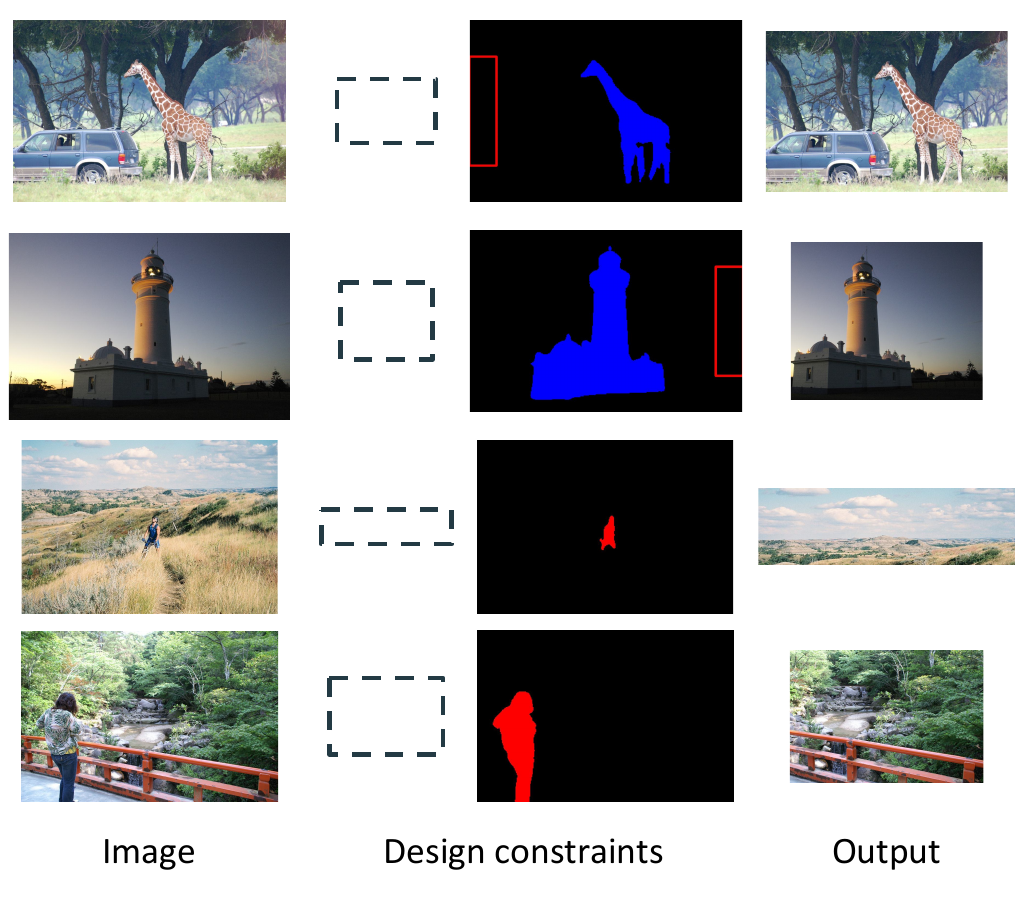}
\vspace{-5mm}
\caption{Examples of application. 
Two upper results show examples of taking in target objects in cropping, while two bottom results show taking target objects off in cropping.}
\label{fig:application}
\end{figure}

\subsection{Application}
The proposed score function-based approach can easily extend the design constraints by adding new score functions.
For example, the proposed approach can control specified objects being in cropping results by replacing a layout box with an object mask.
Also, the proposed approach can take a target object off cropping results by changing the score of the score function to a negative value.
Fig.\ref{fig:application} shows the example of the extension of our method.
\section{Conclusions}
We introduce a novel task image cropping under design constraints and propose a score function-based approach for the task.
As the score function-based approach, we explore the proposal-based approach and the heatmap-based approach.
In experiments, we demonstrate that the proposed approaches outperform simple baselines.
Besides, we observe that the proposal-based approach achieves better performance than the heatmap-based approach under the same computation cost, while the heatmap-based approach achieves better performance with more computation.
This indicates that the balance of obtaining aesthetically plausible and satisfying conditions is sensitive in image cropping under design constraints.
Further, we show the application of image cropping under design constraints.

\clearpage

\bibliographystyle{ACM-Reference-Format}
\bibliography{egbib}

\clearpage
\textbf{\Huge Appendix}

\appendix



\section{Effect of the Hyperparameter $\alpha$}
The hyperparameter $\alpha$ in Eq.~\textcolor{red}{2} balances the score functions of aesthetic and layout conditions.
To investigate the effect of $\alpha$, we conduct a grid search of the hyperparameter $\alpha$ for both the heatmap-based approach and the proposal-based approach.
Table.~\ref{tab:alpha} shows the results of the grid search for the hyperparameter $\alpha$.
We find that leveraging the scores of the layout condition increases the performance of the multiple-conditioned image cropping.
This result indicates that there is a gap in aesthetically plausible areas and the layout conditions.
Note that heavily leveraging the layout score does not mean the aesthetic score is meaningless. It is effective to search aesthetically plausible areas from the areas that satisfy the layout condition in our setting.

\section{Detail of Algorithm}

\paragraph{Aspect ratio specified proposal generation}
We generate aspect ratio fixed proposals in the proposal-based approach.
The pseudo-code for this algorithm is shown in Algorithm.~\ref{alg:proposal}, and we explain the detail of the algorithm to generate aspect ratio fixed proposals.
First, we obtain a base step of the height and width whose ratio matches the given aspect ratio $\omega$ by a grid search (get\_step\_size in Algorithm.~\ref{alg:heatmap}).  Note that we set a minimum value of the base size to 12. If the base size is smaller than 12, we scale up by multiplying a minimum natural value to be over 12.
Second, we linearly increase the base step from $k_{\mathrm{start}}$ to $k_{\mathrm{end}}$.
We set $k_{\mathrm{start}}$ to 14 and $k_{\mathrm{start}}$ to 30, respectively.
Third, we obtain boxes from sliding windows for the box size by setting the minimum size as offsets (sliding\_window in Algorithm~\ref{alg:heatmap}).

\paragraph{Optimization on the heatmap-based approach}

We explore a result of cropping which consists of  ($\mathrm{height},\mathrm{width},\mathrm{position}_{\mathbf{X}},\mathrm{position}_{\mathbf{Y}}$) using the optimization algorithm for a search space.
To address the aspect ratio condition, the search space contains only boxes that satisfy the condition of the aspect ratio in our setting.
If one of the sizes for height or width is determined, the other part is also determined from the aspect ratio. Therefore, in practice, we optimize $(\mathrm{step})$ which is one of size for height or width instead of the pair of $(\mathrm{height},\mathrm{width})$.
In the optimization, we convert a set of parameters ($\mathrm{step}, \mathrm{position}_{\mathbf{X}}, \mathrm{position}_{\mathbf{Y}}$) to a bounding box  $(\mathrm{height},\mathrm{width},\mathrm{position}_{\mathbf{X}},\mathrm{position}_{\mathbf{Y}})$ with given information $(\mathrm{height}_{\mathbf{I}}, \mathrm{width}_{\mathbf{I}}, \omega)$, where $\mathrm{height}_{\mathbf{I}}$ and $\mathrm{width}_{\mathbf{I}}$ are height and width of an image, respectively.
We show the pseudo-code of the conversion in the algorithm~\ref{alg:heatmap}.
Note that we assume that the aspect ratio condition $\omega$ is a scalar value which is obtained by dividing width ratio by height ratio.

\begin{table}[tb]
\centering
\caption{ Grid search of the hyperparameter $\alpha$. The evaluation metric is IoU↑.}
\label{tab:alpha}
\begin{tabular}{l|cc}
\toprule
    $\alpha$      & Heatmap & proposal  \\ \midrule
0.01   & 0.7873 & 0.6447  \\
0.1   & 0.7875 & 0.6753   \\
1   & 0.7875 & 0.7888    \\ 
100  & 0.8013 & 0.8106  \\ 
1e+4  & 0.8283 & 0.8106   \\
\bottomrule
\end{tabular}  
\label{tab:alpha_balance}
\vspace{-3mm}
\end{table}

\begin{algorithm*}
\caption{Aspect ratio specified bounding boxes generation}
\label{alg:proposal}
\begin{algorithmic}[1]
\Function {Proposal}{$\mathrm{height}_{\mathbf{I}}, \mathrm{width}_{\mathbf{I}}, k_{\mathrm{start}}, k_{\mathrm{end}}, \omega$}
    \State{$\mathrm{step}_{\mathbf{H}}, \mathrm{step}_{\mathbf{W}}$ $\leftarrow$ get\_step\_size($\omega$)}
    \State{$\mathcal{Y}_{\omega}^{p} \leftarrow \emptyset	$}
    \For{$k=k_{\mathrm{start}}$ to $k_{\mathrm{end}}$}
        \State{$\mathrm{box}_{\mathbf{H}}\leftarrow$  $k \times \mathrm{step}_{\mathbf{H}}$}
        \State{$\mathrm{box}_{\mathbf{W}}\leftarrow$  $k \times \mathrm{step}_{\mathbf{W}}$}
        \State{$\mathrm{offset}_{\mathbf{H}}\leftarrow$  $\mathrm{step}_{\mathbf{H}}$}
        \State{$\mathrm{offset}_{\mathbf{W}}\leftarrow$  $\mathrm{step}_{\mathbf{W}}$}
        \State{$\mathcal{B} \leftarrow$ sliding\_window($\mathrm{box}_{\mathbf{H}}, \mathrm{box}_{\mathbf{W}}, \mathrm{offset}_{\mathbf{H}}, \mathrm{offset}_{\mathbf{W}}, \mathrm{height}_{\mathbf{I}}, \mathrm{width}_{\mathbf{I}}$)}
        \State{$\mathcal{Y}_{\omega}^{p} \leftarrow$ $\mathcal{Y}_{\omega}^{p} \cup \mathcal{B}$}
    \EndFor
    \State{\textbf{return} $\mathcal{Y}_{\omega}^{p}$}
\EndFunction
\end{algorithmic}
\end{algorithm*}



\begin{algorithm*}
\caption{Conversion for step to height and width in optimization.}
\label{alg:heatmap}
\begin{algorithmic}[1]
\Function {Conversion}{$\mathrm{position}_{\mathbf{X}},\mathrm{position}_{\mathbf{Y}}, \mathrm{step}, \mathrm{height}_{\mathbf{I}}, \mathrm{width}_{\mathbf{I}}, \omega $}
    \State{$\mathrm{margin}_\mathbf{X} \leftarrow \mathrm{width}_{\mathbf{I}} - \mathrm{position}_{\mathbf{X}}$}
    \State{$\mathrm{margin}_\mathbf{Y} \leftarrow \mathrm{height}_{\mathbf{I}} - \mathrm{position}_{\mathbf{Y}}$}
    \State{$\omega_{\mathrm{margin}} \leftarrow \mathrm{margin}_\mathbf{X}/\mathrm{margin}_\mathbf{Y}$}    
    \If{$\omega_{\mathrm{margin}} <= \omega$}
        \State{$\mathrm{height} \leftarrow \mathrm{step}$}
        \State{$\mathrm{width} \leftarrow \mathrm{step} \times \omega$}
    \Else
        \State{$\mathrm{height} \leftarrow \mathrm{step} / \omega$}
        \State{$\mathrm{width} \leftarrow \mathrm{step}$}
    \EndIf
    \State{\textbf{return} $[\mathrm{position}_{\mathbf{X}},\mathrm{position}_{\mathbf{Y}},\mathrm{height}, \mathrm{width}]$}
\EndFunction
\end{algorithmic}
\end{algorithm*}

\section{The effect of hyperparameters in the algorithms}
There are hyperparameters in the algorithm of choice of the best bounding box for cropping.
We decide hyperparameters by grid search.
We show the results of grid search in Table.~\ref{tab:alpha_balance} and Table.~\ref{tab:computation_cost_proposal}.

In the proposal-based approach, we control the number of the proposals by changing $k_{\mathrm{start}}$ and $k_{\mathrm{end}}$ which are the parameters for the size of the offset and the sliding window.  
We show the detail in the Algorithm.~\ref{alg:proposal}.
Table.~\ref{tab:computation_cost_proposal} shows increasing the number of proposal does not guarantee performance improvement in the proposal-based approach. Especially, small $k_{\mathrm{start}}$ brings performance degradation, i.e., increasing the small size of boxes. Note that the size of $k_{\mathrm{end}} = 28$ is a sufficiently large value, i.e., larger $k_{\mathrm{end}}$ does not increase the number of proposal.

The results of Table.~\ref{tab:step_size} shows the effect of the step size in the optimization on the performance of the heatmap-based approach.
We find the size of step affects on the performance of optimization. We expect better step size supports finding better optimal.

\begin{table}[tbh]
    \begin{center}
    \caption{The relationships in the hyperparameters $k_{\mathrm{start}}$ and $k_{\mathrm{start}}$ and IoU score with computaiton cost. \label{tab:computation_cost_proposal}}
    \begin{tabular}{ccccc}
        \toprule
        $k_{\mathrm{start}}$   &   $k_{\mathrm{end}}$ & \% proposals  & IoU↑   & time[s]\\ \midrule
       14   &   16 &  145.59   & 0.3912 & 0.2620\\ 
       14   &   20 &  $\dagger$281.27  & 0.6357 & 0.5052\\ 
       14   &   28 &  $\dagger$303.03   & 0.8106 & 0.5452\\
       12   &   28 &  $\ddag$517.93   & 0.8096 & 0.8822\\
       10   &   28 &  $\ddag$ 801.33   & 0.8088 & 1.4042\\ 
        \bottomrule
    \end{tabular}
    \end{center}
\end{table}

\begin{table}[tbh]
    \begin{center}
   \caption{The relationships in  
   \textit{step size} and IoU score (\textit{iteration}=100). \label{tab:step_size}}
    \begin{tabular}{cc}
        \toprule
        step size & IoU↑   \\ \midrule
         8         & 0.7990 \\
         16        & 0.8128 \\
         32        & \textbf{0.8283} \\
         64        & 0.8250 \\
         128       & 0.7988 \\ \bottomrule
    \end{tabular}
    \end{center}
\end{table}


\section{Detail of Heatmap-based Approach}

\begin{table*}[tb]
\centering
\caption{Model architecture.  }
\begin{tabular}{l|c|c|c}
\toprule
Operation   & input  & output  & hyperparameters \\ \midrule
\multirow{3}{*}{Backbone} &  \multirow{3}{*}{$I$:[256,256,3]} & $F_2$:[64,64,128] & \multirow{3}{*}{-}\\ 
&   & $F_3$:[32,32,256] &\\ 
&   & $F_4$:[16,16,512] &\\ \midrule
Conv + BN + relu &  $F_4$:[16,16,512] & $x$:[16,16,256] & kernel size=3, padding=1\\
Conv + BN + relu &   $x$:[16,16,256] & $x$:[16,16,256]& kernel size=3, padding=1\\
Upsampling &   $x$:[16,16,256] &  $x$:[32,32,256] & scale factor=2\\
Conv + relu &  $x$+$F_3$:[32,32,256]&   $x$:[32,32,128]& kernel size=1, padding=0\\
Upsampling  &   $x$:[32,32,128] &  $x$:[64,64,128] & scale factor=2\\
Conv + relu &  $x$+$F_2$:[64,64,128] &  $x$:[64,64,128]& kernel size=1, padding=0\\
Channel aggregation   & $x$:[64,64,128] &  $x$:[64,64,1] & - \\\bottomrule
\end{tabular}
\label{tab:model_arch}
\end{table*}

\begin{figure*}[t]
\centering
\includegraphics[width=1.0\linewidth]{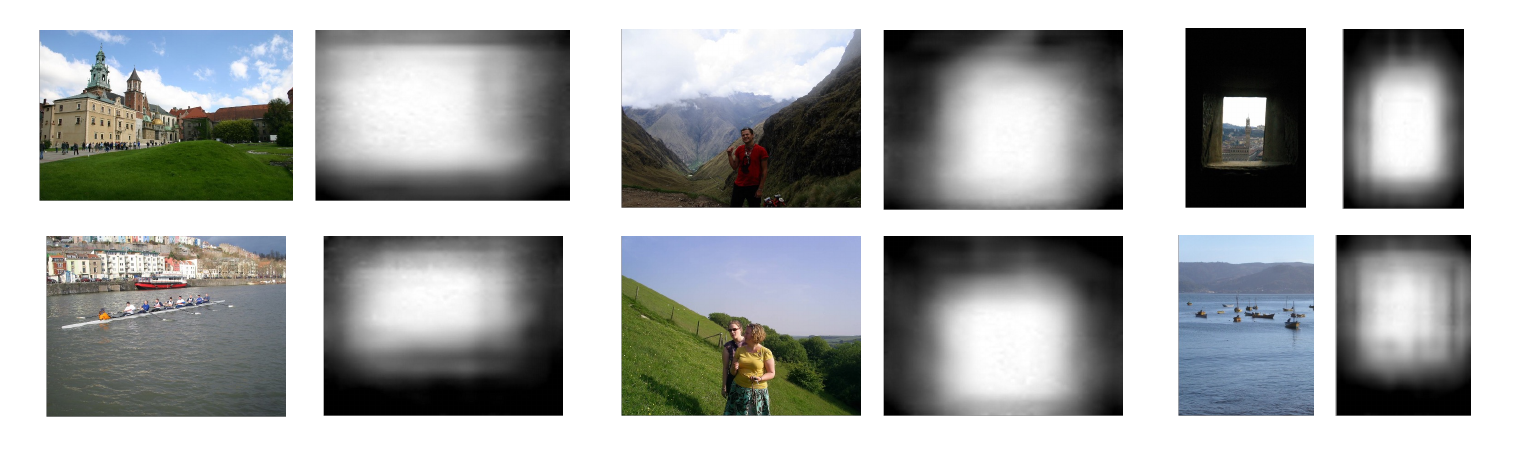}
\vspace{-3mm}
\caption{Examples of heatmaps. }
\label{fig:supplimental_heatmap}
\end{figure*}

\paragraph{Pseudo-heatmap generation for training}
We generate pseudo-heatmaps for the training of the model of the heatmap-based approach.
We compute pseudo-heatmaps following the prior work~\cite{zhang2022human} using the CPC dataset~\cite{wei2018good} and train the model using the pseudo-heatmaps.

\paragraph{Model Architecture}
We show the detail of the architecture of the proposed model in Table.~\ref{tab:model_arch}

\paragraph{Heatmap visualization}
We visualize the heatmaps of the proposed approach in Figure.~\ref{fig:supplimental_heatmap}.
The heatmaps tend to be distributed in the salient area while not responding in the monotonous background area.
Interestingly, the heatmaps include salient objects but the heatmaps contain margin areas. We expect the margin is also important for cropping aesthetically plausible areas.

\section{Limitation}

\paragraph{Computation cost}
Our score function-based approach successfully achieves cropping aesthetically plausible and multiple conditions satisfied areas.
However, the method requires much computation cost for achieving high performance as shown in Table.~\ref{tab:computation_cost_proposal} and Table.~\ref{tab:step_size}.

\begin{figure*}[b]
\centering
\includegraphics[width=1.0\linewidth]{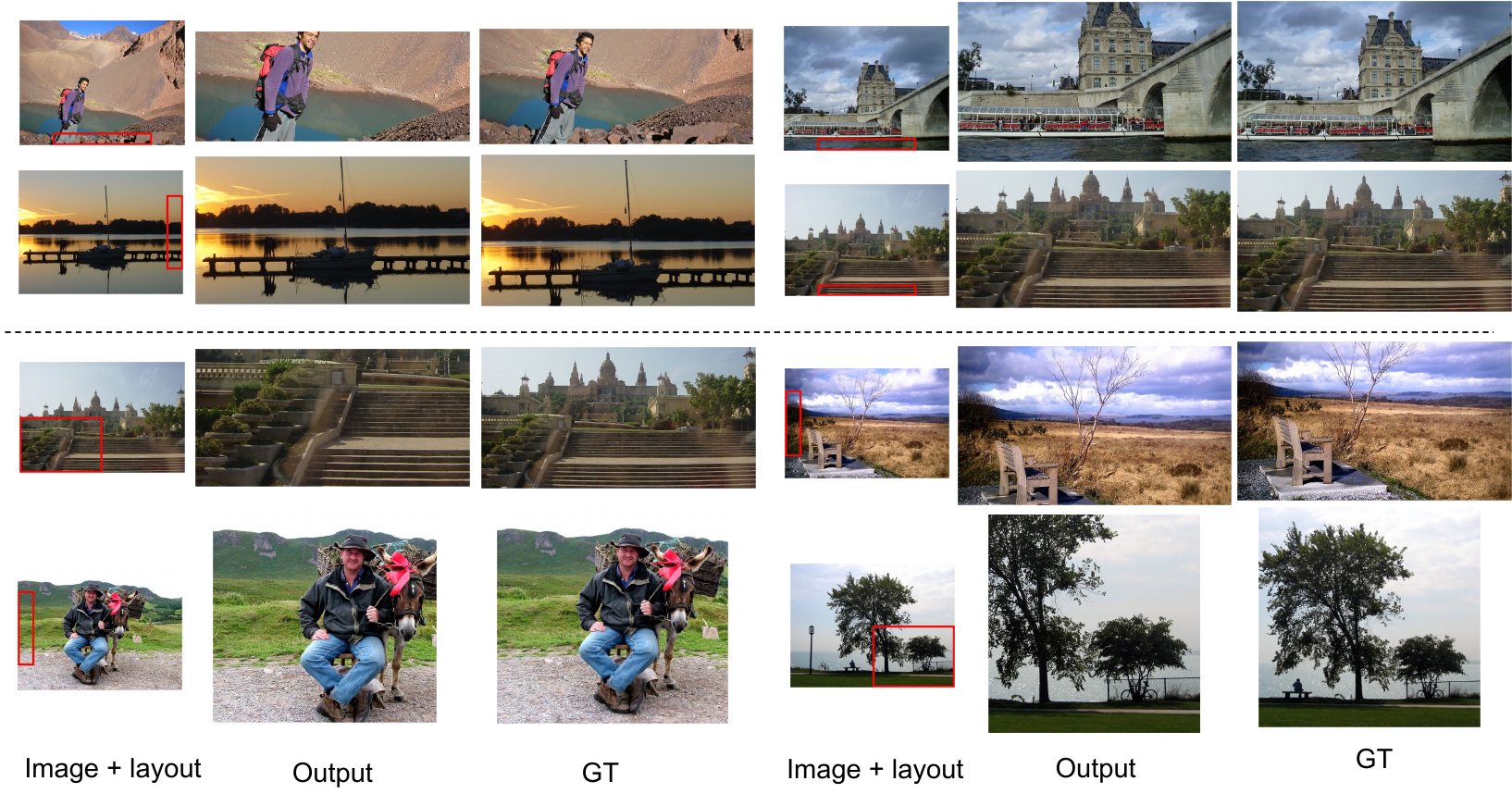}
\caption{This figure shows failure cases of the proposed approaches. The top block shows the failure cases of the heatmap-based approach, and the bottom block shows the failure cases of the proposal-based approach. }
\label{fig:supplimental_lim_heatmap}
\
\end{figure*}

\paragraph{score function of aesthetic}
We explore score functions for aesthetics for heatmaps and scoring of proposals.
Both approaches show good performances but they still contain failure cases.
Figure.~\ref{fig:supplimental_lim_heatmap} shows the failure cases.
The heatmap-based approach tends to fail to contain the full salient object area.
Heatmaps tend to be distributed like Gaussian, and our score function does not sensitive to the edge of objects that are sometimes important for aesthetics.
The proposal-based approach does not have typical failure cases.
We expect the failure cases are from the lack of training data for various aspect ratios.

\paragraph{Templates in evaluation}
Though we use the template layout conditions, eight types
of bounding boxes in evaluation, it is desirable that templates reflect the designers' demand.
To analyze the designers' demand would require large costs, and we use the heuristic templates in this paper.

\paragraph{System in application}
We verify the performance in the situation where layout conditions are given in experiments, but the layout conditions would be variable in the real applications, as we show in Fig.~ref{\ref{fig:application}}.
It is not a trivial problem to define for obtaining the layout condition for specifying regions, and the format and approach should be assumed.
We believe that utilizing detection and segmentation models via interactive interfaces reduces human efforts for specifying the regions.

\section{Additional Qualitative Results}
\begin{figure*}[bt]
\centering

\includegraphics[width=1.0\linewidth]{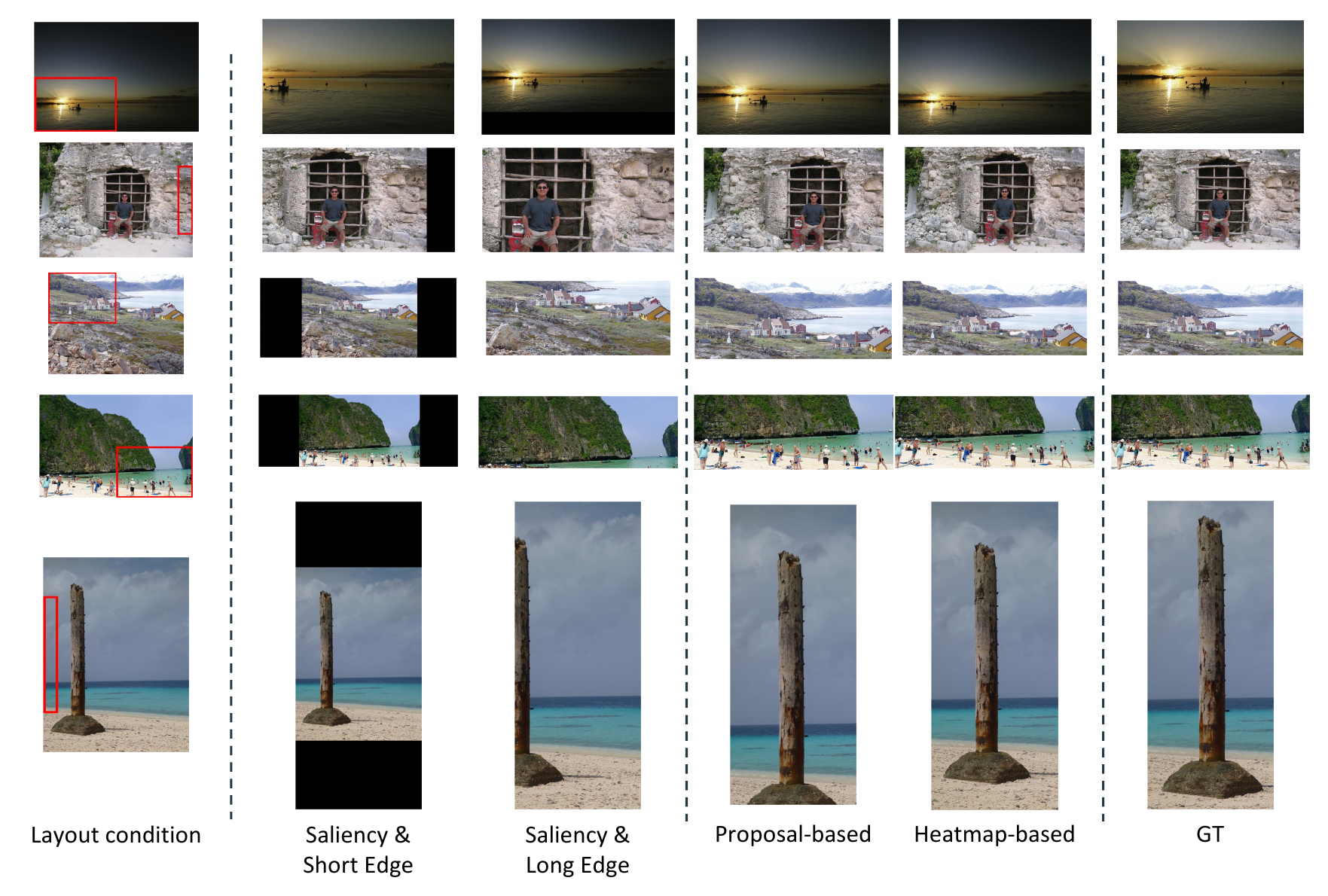}
\caption{This figure shows additional qualitative comparisons. The left image shows an input image
with layout conditions. The middle four images show the multiple conditioned image cropping
by baselines and the proposed approaches. The right figure shows the ground truths.}
\label{fig:supplimental_cmp}
\end{figure*}

\begin{figure*}[t]
\centering
\includegraphics[width=1.0\linewidth]{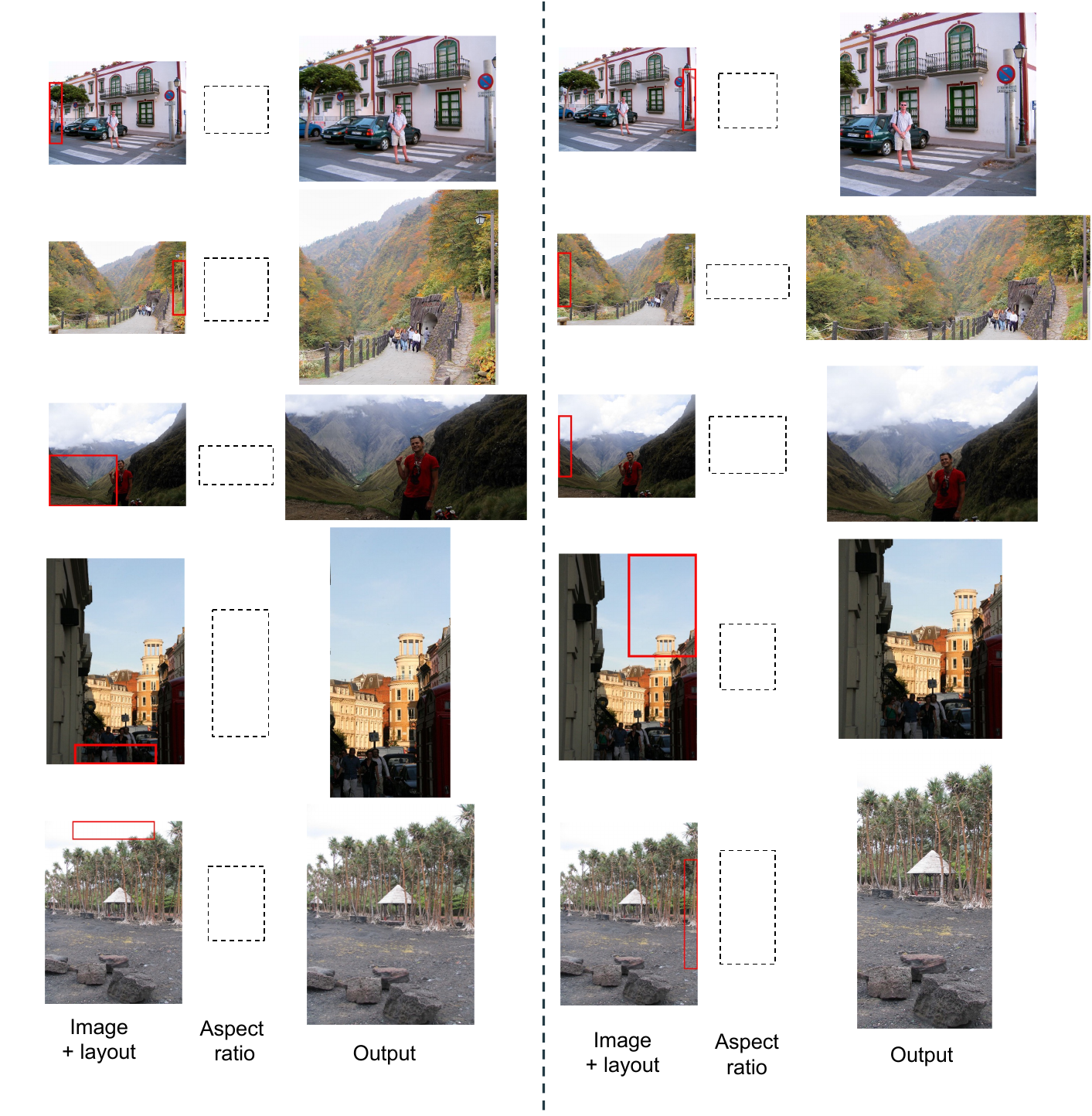}
\caption{Additional examples of the image with multiple conditions and the results of cropping by the
heatmap-based approach.}
\label{fig:supplimental_ex}
\end{figure*}

Figure.~\ref{fig:supplimental_cmp} shows the additional comparisons of the baselines and proposed approaches.
We observe that it is hard to achieve plausible multiple conditioned image cropping by simple baseline approaches, and the proposed approaches successfully obtain the results of cropping that are aesthetically good and satisfy conditions. 
Figure.~\ref{fig:supplimental_ex} shows additional examples of the proposed approach.
The proposed approach can handle various types of aspect ratio and layout specifications.

\end{document}